\documentclass[cameraready]{Interspeech}
\usepackage{makecell}
\usepackage{xcolor}
\usepackage{graphicx}
\usepackage{tikz}
\usetikzlibrary{shapes.geometric, arrows, positioning, calc}
\usepackage{multirow}
\usepackage[colorinlistoftodos]{todonotes}
\usepackage{tipa}
\usepackage{lmodern}

\title{Doing More with Less: Data Augmentation for Sudanese Dialect Automatic Speech Recognition}

\author[affiliation={1}]{Ayman}{Mansour}

\address{
    $^1$ Independent Researcher
}

\email{aymanmnsor777@gmail.com}


\usepackage{comment}

\begin{document}
\maketitle

\begin{abstract}
		Although many Automatic Speech Recognition (ASR) systems have been developed for Modern Standard Arabic (MSA) and Dialectal Arabic (DA), few studies have focused on dialect-specific implementations, particularly for low-resource Arabic dialects such as Sudanese. This paper presents a comprehensive study of data augmentation techniques for fine-tuning OpenAI's Whisper models and establishes the first benchmark for the Sudanese dialect. Two augmentation strategies are investigated: (1) self-training with pseudo-labels generated from unlabeled speech, and (2) TTS-based augmentation using synthetic speech from the Klaam TTS system. The best-performing model, Whisper-Medium fine-tuned with combined self-training and TTS augmentation (28.4 hours), achieves a Word Error Rate (WER) of 57.1\% on the evaluation set and 51.6\% on an out-of-domain holdout set substantially outperforming zero-shot multilingual Whisper (78.8\% WER) and MSA-specialized Arabic models (73.8–123\% WER). All experiments used low-cost resources (Kaggle free tier and Lightning.ai trial), demonstrating that strategic data augmentation can overcome resource limitations for low-resource dialects and provide a practical roadmap for developing ASR systems for low-resource Arabic dialects and other marginalized language varieties.  The models\footnote{\url{https://huggingface.co/collections/AymanMansour/}}, evaluation benchmarks, and reproducible training pipelines are publicly released to facilitate future research on low-resource Arabic ASR.
	\end{abstract}
	
	\section{Introduction}
	Dialectal Arabic includes wide spoken varieties that differ by country and region. While these dialects share key features with Modern Standard Arabic (MSA), their presence in ASR systems varies greatly depending on available data. The Sudanese dialect is underrepresented in Arabic NLP research due to limited training data and few diverse sources for ASR development.
	
	\noindent Major progress in ASR has been made with large multilingual models like self-supervised XLS-R \cite{conneau2020unsupervised} and weakly supervised OpenAI Whisper\cite{radford2023robust}. These pre-trained models, based on huge datasets, allow effective fine-tuning for low-resource languages and dialects, needing much less data than building models from scratch, while data scarcity still remains a challenge for dialects like Sudanese. 
	
	\noindent This work employs self-training, a common semi-supervised method, where a teacher model trained on limited labeled data transcribes unlabeled speech \cite{xu2021self}, and both original and pseudo-labeled data train a student model. Additionally, data augmentation via synthetic speech generation using existing Klaam TTS systems is investigated. 
	
	\noindent The first Sudanese dialect ASR model \cite{mansour2022end} used an end-to-end approach with dilated CNNs and combined MSA/Sudanese data. Building on this, we: fine-tune multiple Whisper versions (Small/Medium/Large V2) on combined MSA-Sudanese data, and explore whether data augmentation (self-training and TTS) can help smaller Whisper models match large model performance. 
	
	\noindent Key contributions include the first comprehensive evaluation and fine-tuning of Whisper models for Sudanese dialect and a new hybrid approach combining self-training with TTS augmentation.
	
	\section{Related work}
	Previous research has shown effective methods for low-resource ASR, with both self-training and synthetic data demonstrating notable promise. Self-training \cite{kahn2020self,xu2021self} has consistently enhanced recognition results across multiple languages, while TTS-generated data has delivered even better outcomes in certain situations, especially when the synthetic speech closely aligns with the target domain \cite{rosenberg2019speech,bartelds2023making}.
	
	\noindent Recent progress in multi-dialect Arabic ASR includes: A multi-dialectal work to apply zero-shot outperform fully fine-tuned XLS-R models, while N-shot has struggled with unseen dialects \cite{talafha2023n}, also a successful fine-tuning of Whisper models for five main dialects (Gulf, Levantine, Iraqi, Egyptian, and Maghrebi), where adding even small amounts of MSA data improved performance \cite{ozyilmaz2025overcoming}. A comprehensive FastConformer-based approach using LLM-generated text augmentation and TTS-synthesized speech data (focused particularly on Egyptian Arabic)\footnote{\url{https://github.com/yousefkorp/Egyptian-Arabic-ASR-and-Diarization}}.

	\section{Datasets}
	
	\subsection{Sudanese Dialect Speech Dataset}
	As the primary resource for the Sudanese dialect \cite{mansour2022end}, it contains 3,544 transcribed recordings (approximately 4 hours) from 31 speakers (9 female, 22 male). The recordings primarily reflect the Khartoum (Central Sudan) dialect with minor Southern influences. Transcriptions were manually generated and do not include diacritics.
	
	\subsection{Arabic Speech Corpus}
	Originally developed for speech synthesis \cite{halabi2016phonetic}, this corpus has been used for multiple ASR systems. It contains 1,813 studio recordings of Formal Arabic spoken with a Southern Levantine (Damascus) accent. Each recording includes:
	Standard transcription, Phonetic transcription, and High-quality studio audio.
	
	\subsection{Lisan-Sudanese TTS Dataset}
	The Novel Sudanese TTS dataset is designed specifically for both ASR and TTS research. Built upon the Lisan-Sudanese Morphological Dataset (52K manually annotated social media tokens from Facebook/X), all sentences were sourced and annotated by Sudanese dialect native speakers \cite{jarrar2023lisan}. the dataset was constructed through a multi-stage pipeline:
	\begin{enumerate}
		\item \textit{Sentence Reconstruction:} by extracting diacritized tokens and reassembling them into full sentences.
		\item \textit{Diacritic Density Filtering:} computed diacritics percentage per sentence using:
		\[
		\text{diacritic\%} = \frac{100}{n}\sum_{t=1}^{n}D_{t}
		\]
		Where: \( n \) = letters + diacritics \( D_{t} \) = \{1 if diacritic, 0 otherwise\}
		\item Retained only sentences with \(\geq\)25\% diacritics (1,878 high-quality examples)
		\item \textit{Synthetic Speech Generation:} Used Klaam\footnote{\url{https://github.com/ARBML/klaam}} TTS (FastSpeech2-based) \cite{ren2020fastspeech} trained on the Arabic Speech Corpus. Generated 4.6 hours of audio with precise text-speech alignment.
	\end{enumerate}
		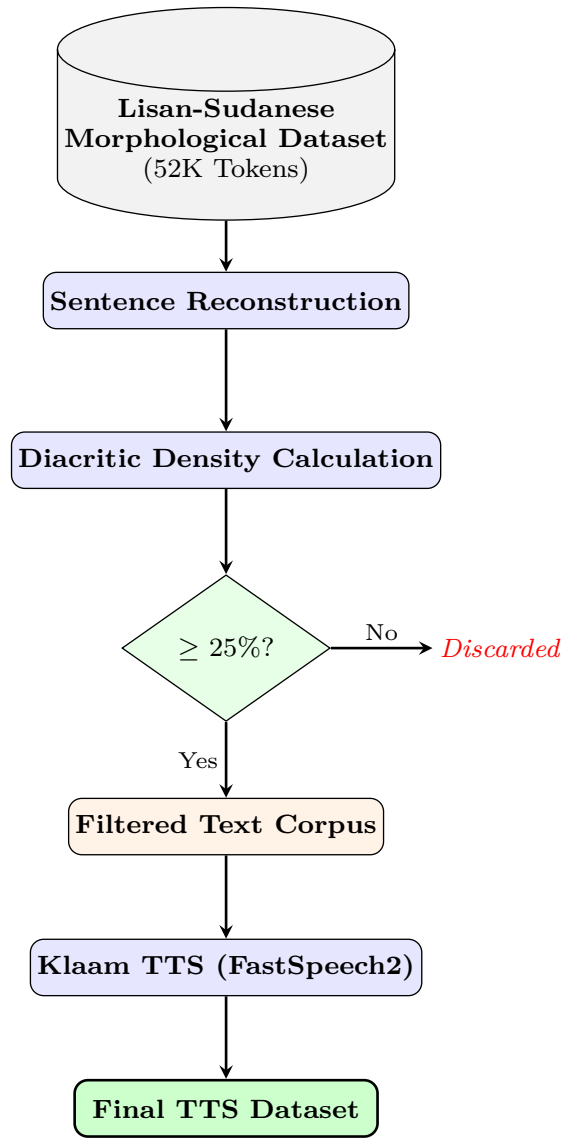
\begin{figure}[t]
		\centering
		\resizebox{0.95\columnwidth}{!}{
			\begin{tikzpicture}[
				node distance=1.5cm, 
				every node/.style={font=\footnotesize, inner sep=2pt}
				]
				\tikzstyle{process} = [rectangle, minimum width=3.2cm, minimum height=0.6cm, text centered, draw=black, fill=blue!10, rounded corners]
				\tikzstyle{database} = [cylinder, shape border rotate=90, draw, minimum height=1cm, minimum width=2cm, shape aspect=0.25, fill=gray!10, text centered]
				\tikzstyle{decision} = [diamond, minimum width=2.2cm, minimum height=0.8cm, text centered, draw=black, fill=green!10]
				\tikzstyle{arrow} = [thick,->,>=stealth]
				
				\node (input) [database, align=center] {\textbf{Lisan-Sudanese}\\ \textbf{Morphological Dataset}\\(52K Tokens)};
				
				\node (recon) [process, below of=input, yshift=-0.2cm] {\textbf{Sentence Reconstruction}};
				\node (filter) [process, below of=recon, yshift=-0.2cm] {\textbf{Diacritic Density Calculation}};
				\node (decide) [decision, below of=filter, yshift=-0.5cm] {$\ge$ 25\%?};
				
				\node (discard) [rectangle, draw=none, right of=decide, xshift=1.4cm, text=red] {\textit{Discarded}};
				
				\node (dataset) [process, below of=decide, yshift=-0.4cm, fill=orange!10] {\textbf{Filtered Text Corpus}};
				\node (tts) [process, below of=dataset] {\textbf{Klaam TTS (FastSpeech2)}};
				\node (output) [process, below of=tts, fill=green!20, thick] {\textbf{Final TTS Dataset}};
				
				\draw [arrow] (input) -- (recon);
				\draw [arrow] (recon) -- (filter);
				\draw [arrow] (filter) -- (decide);
				\draw [arrow] (decide) -- node[anchor=south, font=\scriptsize] {No} (discard);
				\draw [arrow] (decide) -- node[anchor=east, font=\scriptsize] {Yes} (dataset);
				\draw [arrow] (dataset) -- (tts);
				\draw [arrow] (tts) -- (output);
			\end{tikzpicture}
		}
		\caption{Lisan-Sudanese TTS Dataset construction pipeline, detailing the transition from morphological tokens to synthetic audio generation.}
		\label{fig:tts_pipeline_small}
	\end{figure}
	\subsection{OOOK-Eval Dataset}
	Developed by the OOOK\footnote{\url{https://web.archive.org/web/20240619042133/https://www.oook.sd/about}} initiative to expand Sudanese dialect resources, this dataset transcribes the Sudanese portion of ADI17's broadcast collection (48 hours from 30 channels) \cite{shon2020adi17}. The verified subset includes 18,883 raw audio clips, 909 manually transcribed recordings (2.5 hours), annotations by 15 native speakers, and reviewed by a senior annotator, and after performing filtering and pre-processing (\(\sim\)2 hours) of the clean dataset.
	\begin{table*}[h]
		\centering 
		\caption{Dataset Characteristics Summary}
		\label{tab:dataset}
		\begin{tabular}{lclclclcl}
			\toprule
			\multicolumn{1}{c}{\textbf{Dataset}} & \multicolumn{1}{c}{\textbf{Hours}} & \multicolumn{1}{c}{\textbf{Speakers}} & \multicolumn{1}{c}{\textbf{Speaker Overlap}} & \multicolumn{1}{c}{\textbf{Recording Type}} \\ \midrule
			\multirow{1}{*}Sudanese Dialect & 4 & 31 & No& TV broadcasts \\ 
			\multirow{1}{*}Arabic Speech & 3.8 & 1 & No& Studio \\ 
			\multirow{1}{*}OOOK-Eval & \(\sim\)2 & - & Yes& TV broadcasts \\ 
			\multirow{1}{*}OOOK-unlabeled & 45.3 & - & Yes& TV broadcasts \\ 
			\multirow{1}{*}Lisan-Sudanese TTS & 4.6 & 1 & No& Synthetic \\ 
			\toprule
		\end{tabular}
	\end{table*}
	\section{Method}
	A comprehensive experimental framework was adopted to evaluate the performance of state-of-the-art OpenAI's Whisper \cite{radford2023robust} based models on the Sudanese dialect. The method consists of three main stages: zero-shot evaluation, fine-tuning on the manually transcribed Sudanese (SDN) Dialect Speech dataset in combination with MSA, lastly, explore data augmentation (self-training and TTS) by fine-tuning on the generated pseudo-labels and Lisan-Sudanese TTS datasets.
	\subsection{Whisper Models evaluation}
	Three Whisper pre-trained models are evaluated under zero-shot, Whisper\textsubscript{Small}\footnote{\url{https://huggingface.co/openai/whisper-small}}(241M parameters), Whisper\textsubscript{Medium}\footnote{\url{https://huggingface.co/openai/whisper-medium}}(769M parameters), and Whisper\textsubscript{Large-V2} \footnote{\url{https://huggingface.co/openai/whisper-large-v2}}(1550M parameters) are evaluated on OOOK-Eval. Additionally, three Whisper-based models fine-tuned on Arabic and various dialects were evaluated,\textit{arbml/whisper-small-cv-ar}\footnote{\url{https://huggingface.co/arbml/whisper-small-cv-ar}} fine-tuned on Mozilla Common Voice 11 \cite{ardila2020common}, both \textit{arbml/whisper-medium-ar}\footnote{\url{https://huggingface.co/arbml/whisper-medium-ar}} and \textit{arbml/whisper-largev2-ar}\footnote{\url{https://huggingface.co/arbml/whisper-largev2-ar}} are fine-tuned on MGB-2 dataset \cite{ali2016mgb}.
	\subsection{Whisper Models Fine-tuning}
	Three Whisper pre-trained models (Small, Medium, and Large V2) were fine-tuned on a combined dataset of the Sudanese Dialect Speech dataset and the Arabic Speech corpus. 
	Training was conducted for 5,000 steps with a learning rate of 1e-5 and warm-up steps of 500 on a single NVIDIA 40GB A100 GPU. Batch sizes varied by model: (64 training) and (32 evaluation) for Whisper Small, (32 training) and (16 evaluation) for Whisper Medium, and (8 training) and (4 evaluation) for Whisper Large-v2.
	
	\subsection{Self-Training Implementation}
	Due to computational constraints, further experiments focused on Whisper Small and Medium. The self-training approach involved:
	Fine-tuning a teacher model on a cleaned version of the Sudanese dialect Speech dataset, then using this teacher model to generate pseudo-labels for 17,973 unlabeled OOOK recordings, after that applying confidence-based filtering on pseudo-labels predictions with confidence scores for both (\(\geq\) 0.7, \(\geq\) 0.9) to ensure that the model doesn’t reinforce errors, then training a student model on the combined Sudanese dialect dataset and the newly generated pseudo-labeled OOOK dataset. Finally, conducting additional fine-tuning on only generated pseudo-labeled from the OOOK dataset.
	
	\subsection{TTS Data Augmentation}
	Experiments with synthetic data by combining the Sudanese dialect recordings with Lisan-Sudanese TTS-generated speech, Fine-tuning Whisper on this dataset, and also performing fine-tuning on only Lisan-Sudanese TTS. Additional fine-tuning experiments combined both pseudo-labeled and TTS-generated speech.
    
	\noindent \textbf{Self-Training and TTS Data Augmentation fine-tuning Notes:} fine-tuning on whisper Small experiments used Kaggle's free tier (NVIDIA 16GB P100) Batch size reduced to 8 due to GPU memory limits, and whisper medium fine-tuning used Lightning.ai\footnote{\url{https://lightning.ai}} free tier (L40S 48GB). Batch size reduced to 4 and 2 for training and evaluation, respectively. Maintained the same 5,000 steps and 1e-5 and warm-up steps of 500. All implementations used HuggingFace Transformers v4.51.3 \cite{wolf2020transformers}
	
	\subsection{Evaluation}
	
	Performance was reported using the standard ASR metrics, Word Error Rate (\textbf{WER}) shows the percentage of
	word-level errors (insertions, deletions, substitutions) and Character Error Rate (\textbf{CER}) to measure character-level discrepancies.
	
	\noindent{All models were tested using OOOK-Eval to evaluate recognition accuracy and a holdout Sudanese dialect dataset.}
		\begin{table*}[t] 
		\caption{\label{tab:main-result}
			Results Performance Word Error Rate (\textbf{WER}) and Character Error Rate (\textbf{CER}) on OOOK-Eval (/ separated) and \textbf{WER/CER$_{H-O}$} an out-of-domain holdout Sudanese dataset, (con: confidence).} 
		\label{tab:all-results}
		\centering 
		\begin{tabular}{llccc} 
			\toprule
			\textbf{Method} & \textbf{Train Data} & \textbf{Hrs} & \textbf{WER/CER} & \textbf{WER/CER$_{H-O}$} \\
			\midrule
			\multicolumn{5}{l}{\textbf{Zero-shot}} \\
			OpenAI W-Small & Multilingual & 680k & 109/73 & 118/87.8 \\
			OpenAI W-Medium & Multilingual & 680k & 84.3/50.3 & 96.8/65.5 \\
			OpenAI W-Large-V2 & Multilingual & 680k & 78.8/47.7 & 88.5/62.6 \\
			ARBML W-Small-Ar & MGB2 & 1.2k & 83.5/43 & 149/109 \\
			ARBML W-Small-CV-Ar & CV 11 & $\sim$100 & 123/164 & 191/229 \\
			ARBML W-Medium-Ar & MGB2 & 1.2k & 73.8/36.5 & 140/99.6 \\
			ARBML W-Large-v2-Ar & MGB2 & 1.2k & 75.5/37.9 & 146/109 \\
			\midrule
			\multicolumn{5}{l}{\textbf{Full Fine-tuning}} \\
			SDN W-Small & MSA+SDN & 7.25 & 67.7/27.7 & 63.4/32.6 \\
			SDN W-Medium & MSA+SDN & 7.25 & 64.1/26.7 & 57.5/30.9 \\
			SDN W-Large-v2 & MSA+SDN & 7.25 & 62.8/27.1 & 59.7/34.8 \\
			\midrule
			\multicolumn{5}{l}{\textbf{Self-training}} \\
			SDN-Teacher W-Small & SDN-clean & 3.93 & 67.3/26.3 & 61.8/32.2 \\
			SDN W-Small & SDN-clean+Pseudo(con=0.7) & 15.2 & 64.6/25.5 & 60.9/31.3 \\
			SDN W-Small & SDN-clean+Pseudo(con=0.9) & 8.73 & 71/30.5 & 63.5/33.1 \\
			SDN W-Small & Pseudo(con=0.7) & 11.27 & 73.2/29.2 & 64/32.8 \\
			SDN W-Small & Pseudo(con=0.9) & 4.80 & 68.1/25.1 & 68.8/36.2 \\
			SDN-Teacher W-Medium & SDN-clean & 3.93 & 62.5/24.9 & 54/28.3 \\
			SDN W-Medium & SDN-clean+Pseudo(con=0.7) & 23.76 & \textbf{57.1}/\textbf{20.6} & 54.1/28.2 \\
			SDN W-Medium & SDN-clean+Pseudo(con=0.9) & 17.35 & 57.2/20.9 & 52.1/26.6 \\
			SDN W-Medium & Pseudo(con=0.7) & 19.83 & 61/23.7 & 53.3/27.2 \\
			SDN W-Medium & Pseudo(con=0.9) & 13.42 & 58.2/20.9 & 56.5/29.6 \\
			\midrule
			\multicolumn{5}{l}{\textbf{TTS Data-Augmentation}} \\
			SDN W-Small & SDN-clean+TTS & 8.54 & 67.4/27.1 & 61.4/32.1 \\
			SDN W-Small & TTS & 4.61 & 150/118 & 101/69.4 \\
			SDN W-Medium & SDN-clean+TTS & 8.54 & 99.5/74.7 & 57/31.2 \\
			SDN W-Medium & TTS & 4.61 & 71.6/36.3 & 79.9/69.4 \\
			\midrule
			\multicolumn{5}{l}{\textbf{Self-training + Data-Augmentation}} \\
			SDN W-Small & SDN-clean+Pseudo(con=0.7)+TTS & 19.81 & 65.5/26.2 & 64.4/33.6 \\
			SDN W-Small & SDN-clean+Pseudo(con=0.9)+TTS & 13.34 & 70.3/28.1 & 64.5/35 \\
			SDN W-Small & Pseudo(con=0.7)+TTS & 15.88 & 70.5/28.7 & 65.1/33 \\
			SDN W-Small & Pseudo(con=0.9)+TTS & 9.41 & 65.6/25.9 & 63.9/32.3 \\
			SDN W-Medium & SDN-clean+Pseudo(con=0.7)+TTS & 28.37 & 57.9/21.3 & \textbf{51.6}/\textbf{26.5} \\
			SDN W-Medium & SDN-clean+Pseudo(con=0.9)+TTS & 21.96 & 63.3/25.5 & 53.2/28 \\
			SDN W-Medium & Pseudo(con=0.7)+TTS & 24.44 & 61.8/24.1 & 54.6/28.4 \\
			SDN W-Medium & Pseudo(con=0.9)+TTS & 18.03 & 59/21.1 & 52.7/26.7 \\
			\bottomrule 
		\end{tabular}
	\end{table*}
	\section{Results and discussion}
	Table 2 presents a comprehensive evaluation of various Whisper-based ASR models. The analysis reveals key observations across different fine-tuning scenarios: zero-shot evaluation, full fine-tuning, self-training, TTS data-augmentation, and a combinations of both self-training, TTS data-augmentation.
	\subsection{Zero-shot Performance}
	The analysis of zero-shot evaluation showed that when applying out-of-the-box pre-trained models directly on the Sudanese dialect, there is a clear performance gap, although large models are much better at scaling and achieving good results. Whisper {\tiny Large} is by far the best zero-shot model with WER of 78.8\% and CER of 47.7\%, compared to Whisper {\tiny Medium} 84.3\% WER and 50.3\% CER and Whisper {\tiny Small} 109\% WER and 73\% CER on OOOK-Eval. While these results are still high, they indicate that pre-trained models even with 680K of diverse audio, still struggle in capturing the linguistic characteristics of the Sudanese dialect.
	
	\noindent Moreover, interestingly, the ARBML models demonstrated improved performance due to their fine-tuning on both MSA and dialectal datasets, which suggests that Arabic-specific fine-tuning has a slight advantage where \textsubscript{Whisper {\tiny Large-v2-Ar}} achieved 75.5\% WER and 37.9\% CER and \textsubscript{Whisper {\tiny Medium-Ar}} outperformed the Large model by achieving the best WER of 73.8 and CER of 36.5. However, when examining on Sudanese Holdout dataset resulted in a dramatic drop in performance, which implies generalization issues despite their performance, echoing the substantial morphological, phonological, and lexical variations between MSA and regional dialects \cite{habash2010introduction}.
	
	\noindent Finally, the poorest performance across all metrics came from \textsubscript{Whisper {\tiny Small-CV-Ar}}, regardless of Common Voice 100 hours, the model merely scored (WER 123\%, CER 164\%), which suggests not only data quantity, but domain relevance is the paramount factor for ASR performance.
	\subsection{Full Fine-Tuning Performance}
	Full Fine-tuning as targeted adaptation for low-resource scenarios \cite{conneau2020unsupervised} yielded better performance compared to zero-shot baselines, the SDN \textsubscript{Whisper {\tiny Large-V2}} model achieved the best performance with 62.8\% WER and 27.1\% CER on OOOK-Eval. Although the unexpected finding is that SDN \textsubscript{Whisper {\tiny Medium}} performed well compared to the Large model on the holdout set, it scored (57.5\% WER and 30.9\% CER), which demonstrated improved generalization. 
    
	\noindent This observation carries practical implications for low-resource dialect adaptation, suggesting that medium-sized models may offer optimal trade-offs between capacity and data efficiency.
	\subsection{Self-Training Performance}
	The Self-Training emerged as an effective strategy to improve performance for Sudanese dialect ASR. The Medium Teacher model, when trained on the clean Sudanese dialect dataset (3.93 hours), achieved 62.5\% WER on OOOK-Eval and then generated pseudo-labels for unlabeled OOOK audio. Further improvement was reported after a student Medium model is fine-tuned on the combination of SDN-clean and pseudo-labeled{\tiny(con=0.7)} data (23.76 hours total) achieved the best overall performance with 57.1\% WER on OOOK-Eval.
	
	\noindent This represents a relative improvement of approximately 6\% over full fine-tuning Large model, demonstrating the value of leveraging unlabeled Sudanese audio through self-training.
	
	\noindent Whereas, exclusively fine-tuning on pseudo-labeled{\tiny(con=0.9)} data yields moderate performance for the Medium model with (58.2\% WER), indicating that while pseudo-labels contributed with valuable training signal but the importance of gold-standard labeled data is crucial, hence the combination of both pseudo-labeled{\tiny(con=0.9)} and gold-standard demonstrated robust generalization on the holdout dataset with the Medium model achieving 52.1\% WER.
	
	\noindent \textbf{Confidence Threshold Trade-offs}: A confidence threshold of 0.7 outperforms 0.9 in the case of combining pseudo-labeled and gold-standard for the Small model. The lower threshold permits greater data diversity at the cost of potential noise, yet this trade-off proves not beneficial for model robustness when 0.9 outperformed 0.7 with the Medium model performance on holdout set at the same settings.
	
	\noindent \textbf{Model Size Interactions}: Self-training benefits scale disproportionately with model capacity. While the Small model shows modest gains (67.3→64.6 WER), the Medium model demonstrates more pronounced improvements (62.5→57.1 WER), suggesting larger models better leverage noisy pseudo-labels.
	\subsection{TTS Data Augmentation Performance}
	TTS-based data-augmentation presented an interesting result, with the Small model fine-tuned on only TTS-generated data achieving prohibitively high 150\%WER and the Medium model achieving 71.6\% WER on OOOK-Eval.
	However, combining TTS data with gold-standard yielded modest improvement with the Medium model achieving 67.4\% WER, this suggests that synthetic data is linguistically rich (Labels) but failed  due to acoustic distance between TTS output and domain dialect, which manifested in capturing poor variability and naturalness of spontaneous Sudanese dialect speech. 

	\noindent The primary value of TTS augmentation appears to lie in its effect when combined with other techniques rather than as a standalone solution.
	\subsection{Combined Approach Performance}
	The combination of both self-training and TTS data-augmentation alongside gold-standard produced the most compelling results. This combined approach leverages multiple sources: (1) gold-standard provides high-quality acoustic-linguistic mappings, (2) pseudo-labeled data with wide coverage of natural speech variations, and (3) TTS-generated data contributes diverse textual content and potentially improved coverage of vocabulary.
	\noindent Resulting in the best overall generalization across all experimental conditions with SDN \textsubscript{Whisper {\tiny Medium}} fine-tuned on SDN-clean and pseudo-labeled{\tiny(con=0.7)} and TTS data (28.37 hours total) with 51.6\% WER and 26.5\% CER on the holdout dataset, a relative improvement of approximately 45\% over the zero-shot OpenAI\textsubscript{Whisper {\tiny Medium}} baseline (96.8 \%WER/65.5\%CER).

	\subsection{Generalization Gap}
	The performance gap between OOOK-Eval and the holdout metrics is notably obvious; this gap might originate from the speaker overlapping characteristic of each dataset. While OOOK-Eval already has multi-speakers in its recording, the holdout dataset has only a single speaker per clip. Interestingly, Models exhibiting strong OOOK-Eval performance do not uniformly generalize to the holdout set. For instance, ARBML models show competitive OOOK-Eval results but significantly degraded holdout performance. In contrast, fine-tuned and self-trained models maintain more consistent performance across both evaluation settings.

	\subsection{Error Analysis}
\begin{figure} [!ht]
	
	\includegraphics[width=0.45\textwidth]{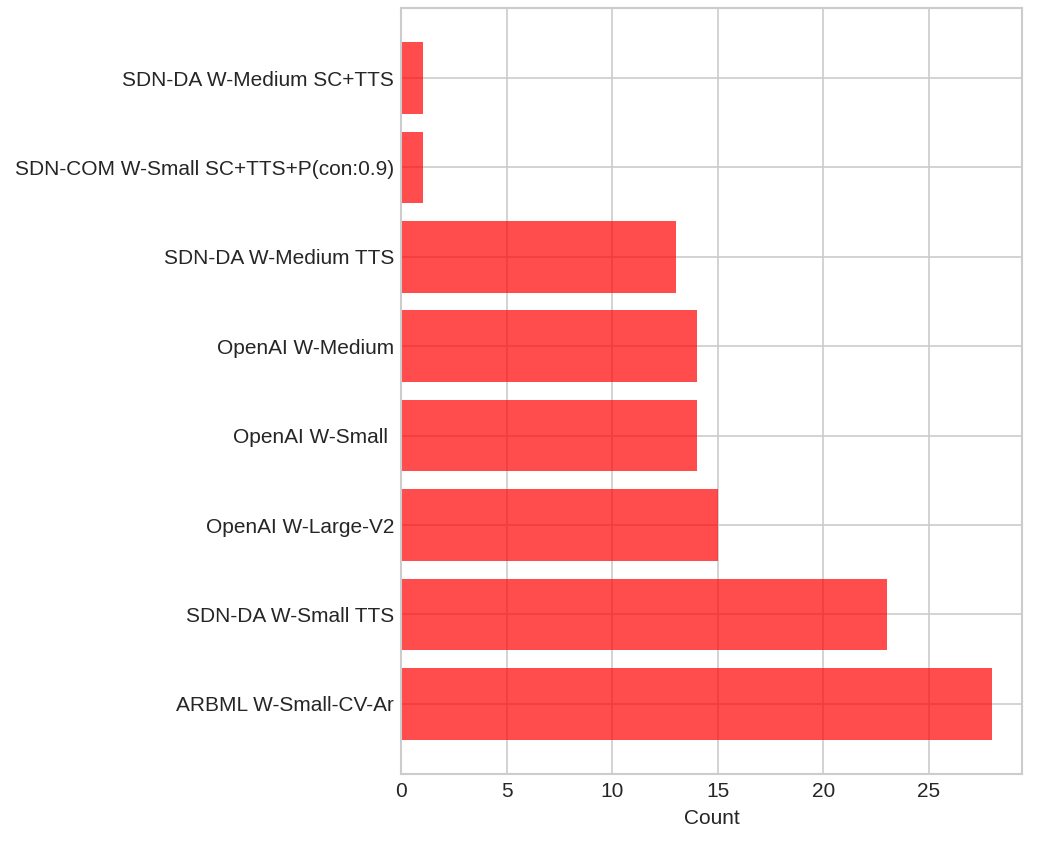}
	\caption{Language Detection Failures}
	\label{fig:LDF}
\end{figure}

\begin{figure} [!ht]
	
	\includegraphics[width=0.45\textwidth]{ 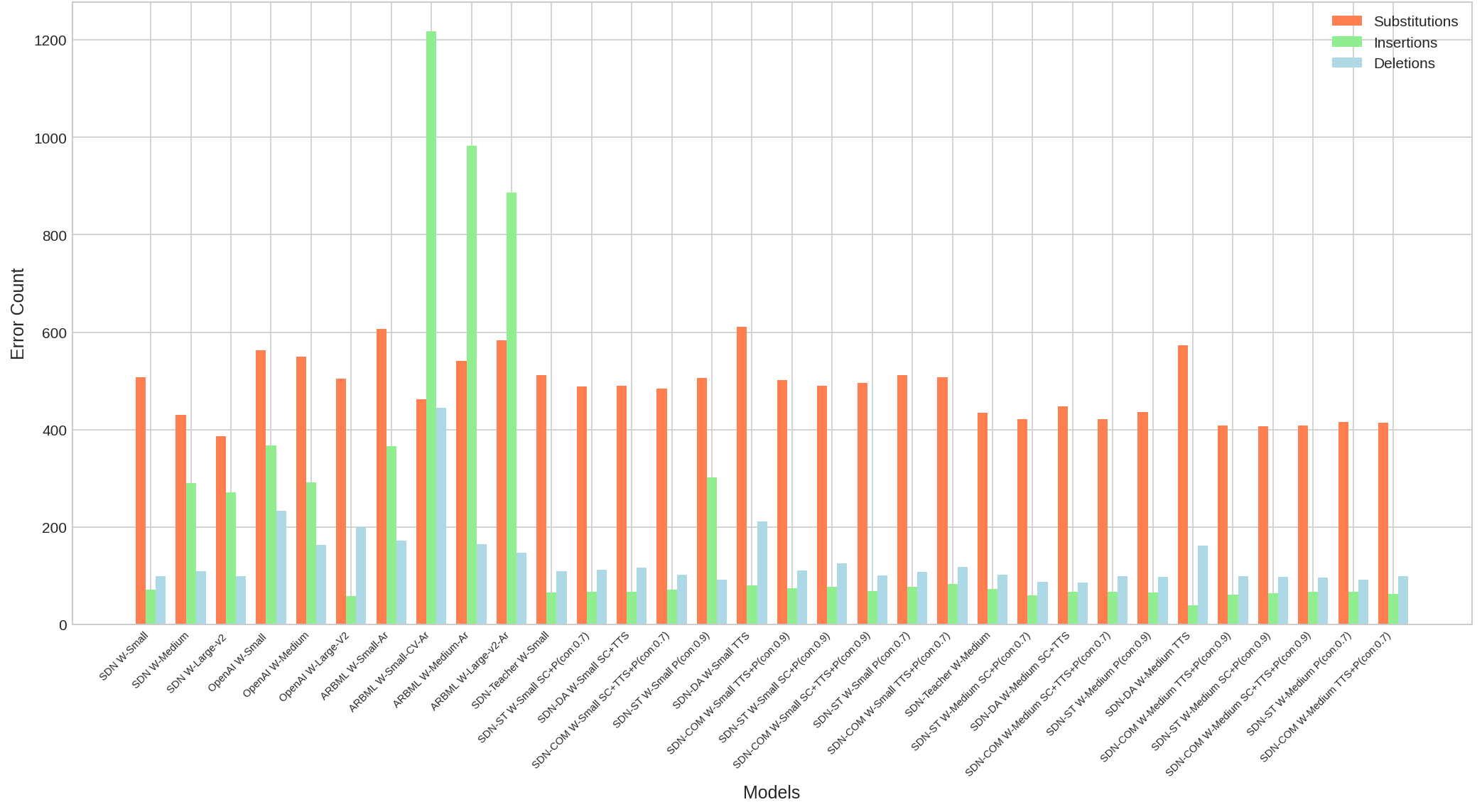}
	\caption{Error Type Distribution}
	\label{fig:ETD}
\end{figure}

\begin{figure} [!ht]
		
		\includegraphics[width=0.45\textwidth]{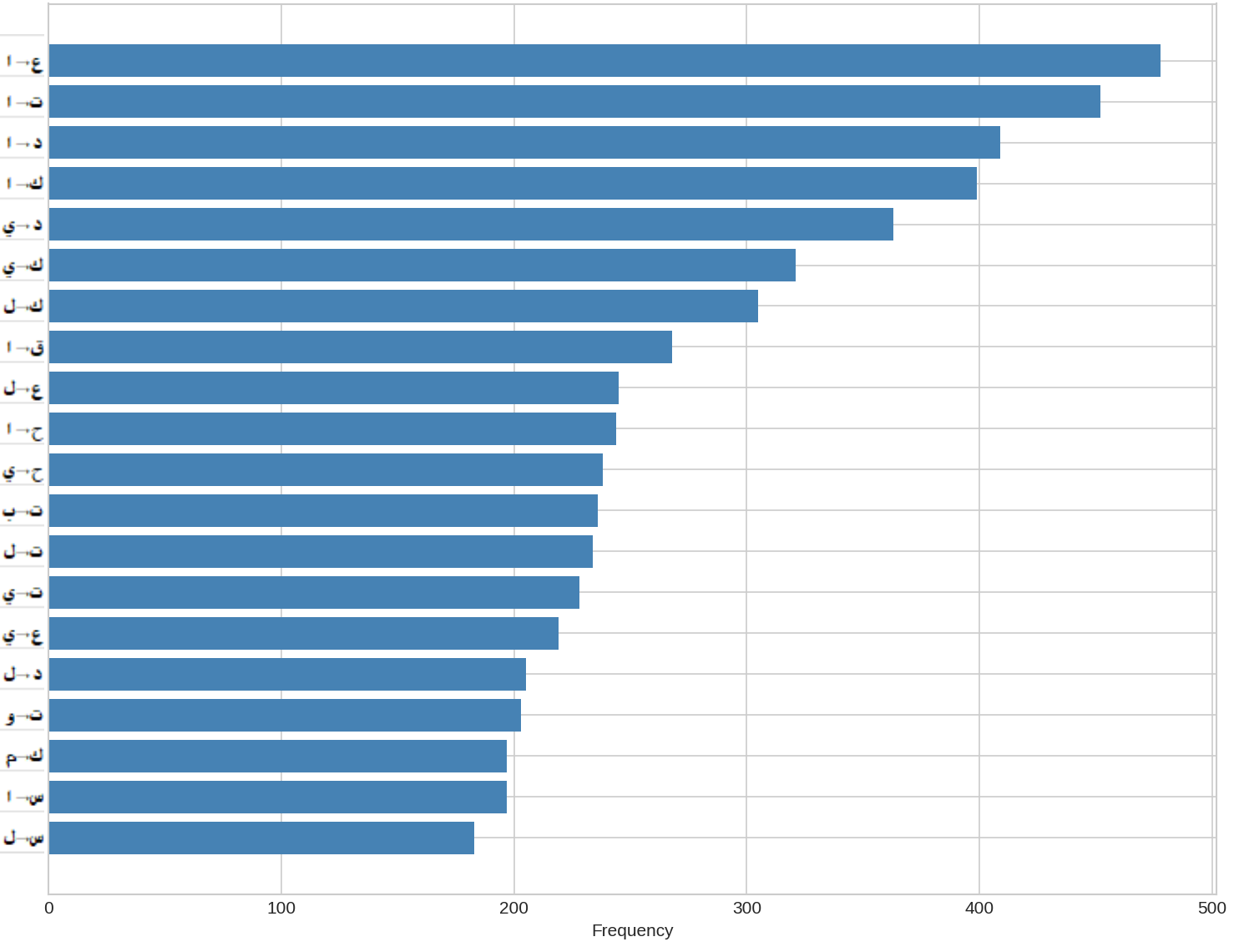}
		\caption{Character Confusions}
		\label{fig:CCL}
	\end{figure}

	A detailed error analysis on 100 samples (holdout) across all models reveals three primary categories of errors: \textbf{Language detection failures:} OpenAI models consistently failed to recognize Sudanese dialect alongside ARBML small model trained exclusively on Common Voice which exhibited the highest failure rates, and notably, models trained solely on synthetic TTS data without clean Sudanese samples(SDN-DA TTS variants) exhibited significant language detection issues (figure~\ref{fig:LDF}). \textbf{Hallucination:} The ARBML W-Small-CV-Ar model exhibited the highest hallucination rate instances, followed by ARBML W-Medium-Ar and ARBML W-Large-v2-Ar (figure~\ref{fig:ETD}). \textbf{Character-level substitution:} (figure~\ref{fig:CCL}) presents the top 20 frequent character-level errors aggregated across all models.

	\section{Conclusion and Future work}
	\noindent This paper presented a comprehensive framework for developing ASR systems for Sudanese dialect. Comparing zero-shot transfer, fine-tuning, self-training, and TTS-based augmentation across various model sizes. The experiments reveal that standard pre-trained models struggle with Sudanese dialect (Zero-shot WER $>$ 73\%), highlighting targeted adaptation for low-resourced dialects. By systematically evaluating adaptation techniques, while direct fine-tuning proves effective, a hybrid approach combining self-training with TTS augmentation offers the most robust performance, achieving a WER of 51.6\% on out-of-domain data, representing a 45\% relative improvement over the zero-shot baseline.
    
	\noindent The analysis yields several key insights for low-resource ASR:
	\begin{itemize}
		\item \textbf{Gold-standard data:} The necessity of clean labeled data is critical for maintaining a foundation for semi-supervised learning.
		\item \textbf{Combined Approach:} Pseudo-labeling and synthetic augmentation are complementary, improving both in-domain accuracy and generalization.
		\item \textbf{Data Efficiency:} diminishing returns of larger model architectures when training data remains limited.
		\item \textbf{Threshold Sensitivity:} Careful calibration of confidence thresholds is essential for balancing accuracy with generalization.
	\end{itemize}
	 While these methods significantly reduced error rates, the best WER above 50\% indicates considerable room for improvement. Future work will explore larger unlabeled corpora, improved TTS systems that better capture dialectal phonetic characteristics, cross-dialect transfer learning, and utilizing parameter-efficient fine-tuning (PEFT) techniques such as LoRA\cite{hu2022lora}.
	 
	 \noindent In conclusion, this work provides both empirical insights and practical methodologies for developing ASR systems for Sudanese Arabic and, by extension, provides a practical roadmap for developing ASR systems for low-resource Arabic dialects and other marginalized language varieties.
	\section{Limitation}
	\noindent This study faces several limitations: While data-augmentation is shown to have clear benefits but the significant gain is dependent on the availability of manually transcribed data. Furthermore,  the Sudanese speech dataset presents limitations regarding size and diversity; it consists of only 4 hours of audio, lacks gender balance (9 female vs. 22 male), and it is geographically restricted to the Central Sudanese dialect. Finally, the TTS system utilizes a single-speaker model based on a Southern Levantine (Damascus) accent, which has notably influenced the performance of the Lisan-Sudanese TTS.
	\section{Ethics Statement}
	\noindent This work evaluated various models and methods aimed at making Automatic Speech Recognition (ASR) systems more viable for the Sudanese dialect and could benefit other under-represented Arabic dialects where paired audio and transcriptions are difficult to obtain. To ensure ethical data usage, only established, open-source datasets were utilized for training and evaluation: The Sudanese dialect speech corpus\cite{mansour2022end}, ADI17\cite{shon2020adi17}, and Lisan-Sudanese\cite{jarrar2023lisan}.

\bibliographystyle{IEEEtran}
\bibliography{mybib}

\end{document}